\documentclass[letterpaper]{article}
\usepackage[preprint]{aaai2027}

\usepackage[hyphens]{url}
\usepackage{graphicx}
\usepackage{natbib}
\usepackage{amsfonts}
\usepackage{caption}
\usepackage{algorithm}
\usepackage{algorithmic}
\usepackage{newfloat}
\usepackage{listings}
\DeclareCaptionStyle{ruled}{labelfont=normalfont,labelsep=colon,strut=off}
\floatstyle{ruled}
\newfloat{listing}{tb}{lst}{}
\floatname{listing}{Listing}

\usepackage{booktabs}
\usepackage{colortbl}
\usepackage{multirow}

\definecolor{AcceptMoEHighlight}{RGB}{229,239,249}
\definecolor{AcceptMoEDelta}{RGB}{27,127,116}
\definecolor{AcceptMoEMuted}{RGB}{110,116,124}
\definecolor{AcceptMoEBand}{RGB}{238,238,240}

\newcommand{\oursrow}{\rowcolor{AcceptMoEHighlight}}
\newcommand{\gain}[1]{\textcolor{AcceptMoEDelta}{#1}}
\newcommand{\delt}[1]{\textcolor{AcceptMoEMuted}{#1}}
\newcommand{\band}[2]{\rowcolor{AcceptMoEBand}\multicolumn{#1}{c}{\textit{#2}}}

\usepackage{xspace}
\newcommand{\sys}{AcceptMoE\xspace}
\newcommand{\sysfix}{AcceptMoE-\ensuremath{B}\xspace}
\newcommand{\moespec}{MoE-Spec\xspace}

\title{AcceptMoE: Commitment-Weighted Self-Sizing Verifier Expert Sets for Efficient MoE Speculative Decoding}

\author{
  Shuang Liang\textsuperscript{\rm 1},
  Hao (Mark) Chen\textsuperscript{\rm 1},
  Zhiwen Mo\textsuperscript{\rm 1},
  Qianzhou Wang\textsuperscript{\rm 1},\\
  Guoyu Li\textsuperscript{\rm 1},
  Lingxiao Ma\textsuperscript{\rm 2},
  Wayne Luk\textsuperscript{\rm 1}
}
\affiliations{
  \textsuperscript{\rm 1}Imperial College London, \textsuperscript{\rm 2}Tile-AI\\
  shuang.liang@imperial.ac.uk
}

\begin{document}

\maketitle

\begin{abstract}
Speculative decoding verifies a tree of draft tokens in one target-model
forward pass. For a mixture-of-experts (MoE) target, however, parallel
verification can activate the union of the experts selected by all tree nodes,
even though only a small subset of those nodes reaches the accepted output.
Token count, activated-expert union size, and expert-weight traffic are
therefore distinct cost measures: reducing the token workload need not shrink
the expert union proportionally, and under offloading, transfer traffic also
depends on cache residency.
We introduce \sys, a verifier-side expert selector that combines target-router
scores with offline-estimated commitment probabilities and automatically
adjusts the number of eligible experts for each verification block, eliminating
the need for a user-specified expert budget. Under
offloading, \sys conditions expert eligibility on cache residency instead of
predicting natural routes and prefetching the corresponding expert weights.
Although constraining target-expert eligibility changes the model distribution,
across 12 model--task pairs spanning three MoE targets and four benchmarks,
\sys's mean accuracy is $0.27$ percentage points lower than that of EAGLE-3
speculative decoding with natural routing. Served with SGLang at batch size
one, it reaches $1.290\times$ the throughput of this baseline with all expert
weights in GPU memory, and $2.06\times$ under physical expert offloading,
while reducing host-to-device traffic by $73.6\%$ to $77.1\%$.
\end{abstract}

\section{Introduction}

Autoregressive decoding remains a dominant cost in large language model (LLM)
inference because every output token requires a full forward pass through the
model~\citep{yang2025qwen3, guo2025deepseek, agarwal2025gpt}.
Speculative decoding~\cite{cai2024medusa, li2024eagle, chen2024hardware}
amortizes this cost by using a draft mechanism to propose a block of tokens
that the target verifies in one forward pass. The verifier then accepts a
prefix according to a rule that preserves the target distribution.

For an MoE target, however, parallel verification need not reduce memory
traffic proportionally. Although each token activates only its top-$k$
experts, different draft-tree nodes can select different
experts~\cite{dai2024deepseekmoe}. A verification block can therefore access
the union of all per-node top-$k$ expert sets, whose cardinality may
substantially exceed $k$. The size of this union determines the number of
distinct expert weights accessed during verification. \textbf{Challenge 1:
Natural routing incurs expert-weight
traffic for draft branches that are ultimately discarded, so reducing the
number of verified tokens need not proportionally reduce the activated-expert
union.}

\emph{Draft-side selection} determines which draft nodes are retained for
target verification, whereas \emph{verifier-side expert selection} determines
which target experts remain eligible for the resulting verification block. The
two decisions operate on different quantities. Reducing the number of retained
draft nodes need not shrink the activated-expert union proportionally. On
Qwen3-30B-A3B, EVICT reports $74.7\%$ fewer verified tokens
than EAGLE-3 but only $32.5\%$ fewer activated experts~\citep{pan2026making}.

A verifier-side expert selector determines the eligible expert set to which
each token's routing is restricted. \moespec constructs this set by aggregating
router mass over the verification block and retaining a prescribed number
$B$ of experts at each layer~\cite{mcdanel2026moe}. Its set size is therefore
externally specified, while its aggregation weights every draft position
equally despite their unequal probabilities of reaching the output.
\textbf{Challenge 2: A fixed expert budget must be supplied in advance, but
the best measured budget varies across evaluated model--task pairs.}

\sys addresses both limitations by constructing commitment-weighted router
demand for each verification block. For each draft position, we estimate
offline the probability that a node proposed there is committed to the output
and use this probability to weight target-router scores before summing them
across the block. \sys ranks experts by this weighted demand and retains the
highest-ranked non-anchor experts while always preserving the natural top-$k$
experts of the root token. It determines how many non-anchor experts to retain
using the effective rank of their demand distribution, defined as the
exponential of its entropy. Concentrated demand yields a smaller set, whereas
diffuse demand yields a larger one. The resulting expert set is therefore
recomputed and self-sized for every verification block without a predefined
expert budget. Across the 12 model--task pairs, the rounded per-pair mean set
size selected by \sys ranges from $23$ to $34$ experts.

\textbf{Challenge 3: Under expert offloading, transfer cost depends on which
selected experts are nonresident, not only on the set cardinality.} \sys
therefore applies \emph{residency-aware pruning}: it orders selected
nonresident experts by increasing commitment-weighted demand and removes the
longest prefix allowed by the rerouting budget and minimum-set-size constraint.
Prior offloading systems for speculative MoE decoding predict the target's
natural expert routes and prefetch the corresponding expert weights. \sys
instead makes expert eligibility a function of the resident set, without
requiring a learned expert predictor or a predefined expert budget. Our contributions are as follows.
\begin{itemize}
    \item \textbf{Commitment-weighted expert demand.} To resolve
    \textbf{Challenge~1}, we replace the uniform router-mass aggregation of prior
    verifier-side selectors with target-router scores weighted by
    position-dependent commitment probabilities estimated offline. This places
    greater weight on demand from draft positions that are more likely to be
    committed. At matched per-pair budgets $B_0$, \sysfix improves the
    mean accuracy across the 12 model--task pairs by $2.45$ percentage points
    over \moespec.
    \item \textbf{Self-sizing expert sets.} To resolve \textbf{Challenge~2},
    we derive the expert-set size from the effective rank of the same demand,
    removing the expert budget that prior selectors must be given in advance.
    Across five budget-swept pairs, the self-sized point is $0.97$ percentage
    points below the best measured fixed-budget point on average and at most
    $1.83$ points below it.
    \item \textbf{Residency-aware pruning.} To resolve \textbf{Challenge~3},
    we evaluate the same demand against the resident set and prune low-demand
    nonresident members, making expert eligibility a function of the expert
    cache rather than a prefetch schedule. Compared with the same selector with
    pruning disabled, this stage reduces expert-weight traffic by $38.6\%$ to
    $48.6\%$.
    Its mean accuracy change over the 12 pairs is $-0.27$ percentage
    points, with per-pair changes from $-1.83$ to $+1.22$ points.
\end{itemize}

Measured end to end in the SGLang serving stack across three MoE targets
and four tasks at batch size one, \sys's mean accuracy is \textbf{0.27}
percentage points lower than that of Standard SD, EAGLE-3 speculative decoding
with the target's natural routing.
It reaches \textbf{1.290}$\times$ the throughput of Standard SD with all
experts resident, decoding at up to \textbf{257} tokens per second on one RTX
PRO 6000 Blackwell, and \textbf{2.06}$\times$ under expert offloading on one RTX 5090,
while reducing host-to-device traffic by \textbf{73.6\% to 77.1\%}.

\section{Related Work}

\begin{figure*}[t]
\centering
\includegraphics[width=\textwidth]{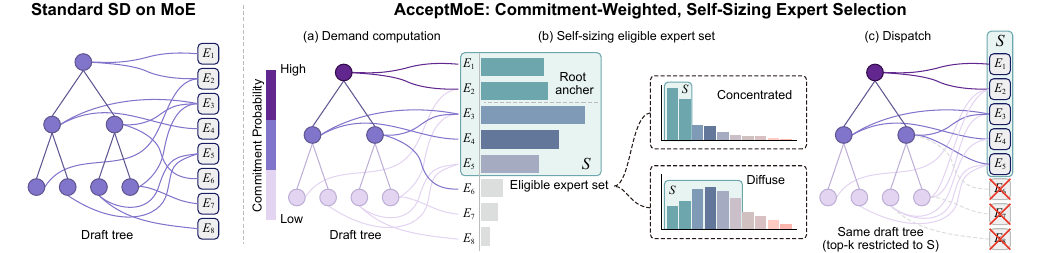}
\caption{\sys constructs a layer- and block-specific verifier expert set.
Position-dependent commitment probabilities weight target-router demand, the
root token's natural top-$k$ experts seed the set, and effective rank determines
how many additional experts are retained. Figure~\ref{fig:algorithm_part2}
shows the residency-aware pruning applied when expert weights are offloaded.}
\label{fig:algorithm_part1}
\end{figure*}

\subsection{Speculative Decoding}
Speculative decoding uses a draft mechanism to propose a block of tokens that
the target model verifies in one parallel forward pass,
with an accept/reject rule that preserves the target
distribution~\citep{stern2018blockwise,leviathan2023fast}. Many subsequent methods vary
the drafter. Medusa and PPD let the target draft its own continuations through
parallel decoding heads or trained prompt tokens~\citep{cai2024medusa,chen2024hardware}. EAGLE
drafts autoregressively in feature space, with EAGLE-2 adding dynamic draft
trees and EAGLE-3 using direct token drafting over fused multi-level
features~\citep{li2024eagle,li2024eagle2,li2026eagle3}.

\subsection{Mixture-of-Experts Optimization}
Sparse MoE layers decouple parameter count from per-token compute and are used
by many large open-weight models~\citep{jiang2024mixtral,dai2024deepseekmoe}.
Their memory footprint has motivated \emph{expert pruning}, which removes
experts offline from router or task
statistics~\citep{chen2022task,zhang2025diversifying}, and \emph{expert
skipping}, which drops low-value experts per token at run
time~\citep{huang2025modes,lu2024not,huang2024mixture,aghdam2024moe}.
\emph{Expert sharing} instead routes a parallel workload through a shared
expert subset. Opportunistic activation reuses experts already selected in the
batch~\citep{oncescu2025opportunistic}, XShare maximizes aggregate gating mass
over a per-batch subset~\citep{vankov2026xshare}, and dynamic expert sharing
selects a compact expert coreset for a parallel diffusion-decoding
block~\citep{chen2026dynamic}. Expert sharing is the family closest to ours,
since it also chooses one expert set for a parallel workload. \sys specifically
uses the position-dependent commitment structure of a speculative verification
tree and derives the expert-set cardinality for each block.

\subsection{Speculative Decoding for MoE Models}
A recent line of work studies how speculation interacts with MoE sparsity.
MoESD analyzes how batch size and sparsity affect target efficiency and
speculative speedup~\citep{huang2025moesd}, whereas Cascade measures speculation
utility online and selects a speculation length, disabling speculation
altogether when its measured utility falls below
one~\citep{saxena2025utility}. Neither method changes verifier expert routing.

Under expert offloading, SP-MoE uses draft-model attention outputs with the
target router to predict and prefetch experts needed for
verification~\citep{chen2025spmoe}. MoE-SpeQ uses a quantized MoE draft model
for multi-step expert lookahead, then orchestrates prefetching, caching, and
draft length~\citep{wang2025moespeq}. Both systems predict the target's natural expert
routes and schedule the resulting weight movement. Neither changes which
experts remain eligible during verification.

Draft-side selection determines which draft nodes enter target verification.
EVICT selects an ancestor-closed prefix by combining drafter scores with
offline-profiled verification cost~\citep{pan2026making}. EcoSpec addresses
expert scattering by fine-tuning a target-specific expert predictor and using
a dynamic expert buffer to select a specified number of draft nodes based on
acceptance likelihood and predicted marginal expert
cost~\citep{xie2026ecospec}. Both methods preserve the target's natural
top-$k$ routing and the distribution-preserving semantics of speculative
verification while selecting the retained draft-node set before target-model
verification begins. In contrast, \sys runs inside the serving engine after the
verification block is fixed and requires no auxiliary expert predictor, routing
supervision, or additional model inference.

Verifier-side methods instead constrain expert eligibility after the
verification block has been fixed. In its speculative mode, XShare performs
hierarchical expert selection under per-request and batch
capacities~\citep{vankov2026xshare}. \moespec aggregates router probabilities
over the draft tree and retains the top $B$ experts at each
layer~\citep{mcdanel2026moe}. These methods select experts subject to
externally specified capacity constraints. \sys instead derives both expert-set
membership and per-block cardinality from commitment-weighted demand and, under
offloading, prunes nonresident members based on the current resident set.


\section{Method}
\label{sec:method}

\sys constructs one eligible expert set for each MoE layer and verification
block. Figure~\ref{fig:algorithm_part1} summarizes the three stages used under
full residency. The first panel scores each expert by
\emph{commitment-weighted demand}, obtained by weighting the target router's
natural top-$k$ probabilities by the marginal probability that a node at the
same draft position is committed to the output. The second panel combines a
root anchor with this demand. Demand ranking selects the non-anchor experts,
and demand concentration determines how many are retained. The final panel
restricts every token's routing to the resulting set during verification.
Under expert offloading, Figure~\ref{fig:algorithm_part2} shows the additional
residency-aware pruning stage, which removes low-demand nonresident experts
subject to the rerouting budget and minimum set size.

\subsection{Verifier-Side Expert Selection}
\label{sec:method_formulation}

Consider one target MoE layer with $E$ experts and per-token top-$k$ routing.
A draft-side selection policy has already produced a verification block
$\mathcal V$ containing $T$ tokens. For token $t\in\mathcal V$, let
$z_t\in\mathbb{R}^{E}$ denote the target-router logits,
$r_t=\mathrm{softmax}(z_t)$ the router distribution, and $d_t$ the draft
position, given by block offset for a chain and tree depth for a tree.
Natural routing selects
$K_t=\mathrm{TopK}_{k}(r_t)$, so the layer's activated expert union is
\begin{equation}
\label{eq:natural_union}
U_{\mathrm{nat}}(\mathcal V)=\bigcup_{t\in\mathcal V}K_t.
\end{equation}
A verifier-side expert selector constructs one layer- and block-specific
eligible expert set $S\subseteq[E]$, where $[E]=\{1,\ldots,E\}$. The layer routes each
token by
\begin{equation}
\label{eq:constrained_routing}
\widetilde K_t(S)=\mathrm{TopK}_{k}\!\left(z_t+M_S\right),
\end{equation}
where $(M_S)_e=0$ for $e\in S$ and $(M_S)_e=-\infty$ otherwise. The
implementation applies this logit mask, or an equivalent zero-probability mask,
in the corresponding backend and preserves the target's native post-top-$k$
weighting. Dispatch under this constraint is stage (c) of
Figure~\ref{fig:algorithm_part1}, and setting $S=[E]$ recovers Standard SD. A
fixed-budget selector takes $|S|$ as a hyperparameter and determines only
membership, whereas \sys derives both membership and cardinality from the
current block. Because the mask
changes the target router's logits, any such selector is an approximation
rather than a distribution-preserving optimization.

\subsection{Commitment-Weighted Demand}
\label{sec:method_utility}

Aggregate router mass, $\sum_t r_{t,e}$, weights every token in the verification
block equally. Two properties of the draft tree argue against this choice.
Deeper positions are less likely to reach the accepted prefix, and sibling
branches are mutually exclusive because at most one root-to-leaf path can be
committed; consequently, their demands are never realized together. \sys therefore weights each position
by how often a node proposed there is committed to the model's output. For draft position $d$, we estimate
$\widehat p_d=c_d/m_d$ from offline verification traces, where $m_d$ is the
number of proposed nodes observed at position $d$ and $c_d$ is the number of
those nodes that lie on the accepted output prefix. This is a marginal
commitment probability, not the conditional acceptance rate usually reported
for speculative decoders. The estimates are computed from training-split
traces, fixed before evaluation, and never use evaluation labels.

Let $N_d=\sum_t\mathbf{1}[d_t=d]$ be the number of tokens at position $d$ in
the block. \sys assigns token $t$ the importance
\begin{equation}
\label{eq:token_importance}
\alpha_t=\widehat p_{d_t}^{\,\beta}N_{d_t}^{-1/2},
\end{equation}
where $\beta$ controls how sharply the commitment estimates are applied
(default $\beta=0.5$). The $N_{d_t}^{-1/2}$ factor discounts wide tree levels
sublinearly, so their aggregate weight grows as $\sqrt{N_d}$ rather than
$N_d$. Expert utility is
\begin{equation}
\label{eq:aware_utility}
u_e=
\sum_{t=1}^{T}\alpha_t r_{t,e}\mathbf{1}[e\in K_t],
\qquad e\in[E].
\end{equation}
An expert therefore receives router probability mass only through the natural
top-$k$ set of some token, with lower weights at positions that are less likely
to be committed. Because Eq.~\ref{eq:aware_utility} uses only target-router
outputs already available in the verifier and the offline position weights, it
requires no additional model inference.

\subsection{Self-Sizing the Eligible Expert Set}
\label{sec:method_select}

Expert-set construction ranks experts by $u_e$ and always includes the root
anchor
\begin{equation}
A=\bigcup_{t\in\mathcal T_{\mathrm{anc}}}K_t,
\end{equation}
where $\mathcal T_{\mathrm{anc}}$ contains the designated anchor tokens. For
EAGLE this is the depth-0 root. Before residency-aware pruning, its natural
top-$k$ set guarantees coverage for the one token that is certain to be
committed, while every other eligible expert is selected by estimated demand.

Let $R=[E]\setminus A$ denote the non-anchor experts and $U_R$ the utilities
restricted to $R$. \sys retains the top-ranked experts of $R$ and derives
their number from the shape of $U_R$. Let
$u_e^+=\max(u_e,0)$ for $e\in R$ and
$Z=\sum_{j\in R}u_j^+$. When $Z>0$, define $q_e=u_e^+/Z$ and
\begin{equation}
\label{eq:effrank_cardinality}
n_{\mathrm{er}}=
\left\lceil\exp\!\left(-\sum_{e\in R}q_e\log q_e\right)\right\rceil.
\end{equation}
The eligible expert set is $S=A\cup\mathrm{Top}_{n_{\mathrm{er}}}(U_R)$, of
total size $|A|+n_{\mathrm{er}}$. The quantity $n_{\mathrm{er}}$ is the
effective-rank functional~\citep{roy2007effective} applied to the demand
distribution, namely the exponential of its entropy. It
approaches one when residual demand concentrates on a single expert and $m$
when the demand spreads uniformly over $m$ experts, so a block whose demand is
concentrated is served by a smaller set. A minimum-size constraint enforces
$|S|\ge k$. If the residual utilities are all zero, \sys retains only as many
non-anchor experts as that constraint requires.

\subsection{Residency-Aware Pruning}
\label{sec:method_residency}

\begin{figure}[tb]
\centering
\includegraphics[width=\columnwidth]{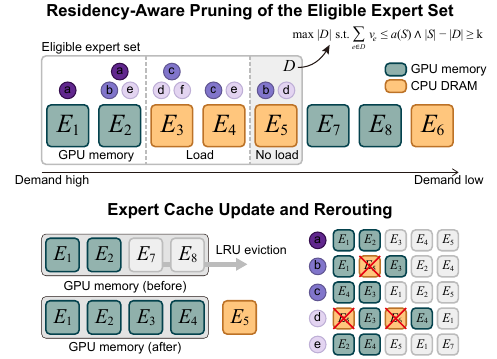}
\caption{Residency-aware pruning of the \sys expert set. All nonresident
members, including experts contributed by the anchor, are ordered by
commitment-weighted demand; the rerouting budget and the minimum set size
determine how long a prefix is removed.}
\label{fig:algorithm_part2}
\end{figure}

The size of $S$ bounds the number of distinct experts that can be activated,
but does not determine weight-transfer traffic under partial residency. A
resident expert executes without a host-to-device transfer, whereas a
nonresident expert triggers a weight load and can evict a cached expert. \sys
therefore treats the resident set as a runtime signal of transfer cost and
prunes $S$ with the same utility, as illustrated in
Figure~\ref{fig:algorithm_part2}.

Each MoE layer has an $X$-slot expert pool in GPU memory and holds the full
expert weights in pinned host memory. The pool caps residency, not
eligibility. A nonresident expert is loaded synchronously when it receives
tokens. If the block's routed expert union after masking exceeds $X$, tokens
are partitioned into several waves, each of whose routed expert union fits in
the pool.

For each layer, let $\mathcal H$ be the current LRU resident set and
$C=S\setminus\mathcal H$ the nonresident members of $S$. \sys removes a subset
$D\subseteq C$ and routes the block through $S\setminus D$ under the same
constrained top-$k$ rule as Eq.~\ref{eq:constrained_routing}. Tokens whose
natural experts were removed are rerouted to retained experts, which need not
themselves be resident. Any such expert is loaded on demand when it first
receives a token. At full residency $C=\emptyset$ and this step leaves $S$
unchanged.

The removal is bounded by the rerouting that forming $S$ has already caused.
Let $v_e=\sum_t\mathbf{1}[e\in K_t]$ be the token assignment count under
natural top-$k$ routing and $K=\bigcup_tK_t$. Restricting the layer to $S$
already displaces $a(S)=\sum_{e\in K\setminus S}v_e$ natural assignments, and
we allow pruning to displace at most that many more, which we call the
rerouting budget. Let $\pi$ order the experts in $C$ by increasing utility
$u_e$, breaking ties by expert index. We remove the longest prefix that
respects the rerouting budget and the minimum-size constraint:
\begin{equation}
\label{eq:residency_objective}
\begin{array}{l}
m^\star=\max\bigl\{m:\sum_{j=1}^{m}v_{\pi(j)}\le a(S),\\
\qquad\qquad |S|-m\ge k\bigr\},\\
D^\star=\{\pi(1),\ldots,\pi(m^\star)\}.
\end{array}
\end{equation}
The final eligible expert set is $S'=S\setminus D^\star$. In the evaluated
configuration, the implementation's optional removal cap is set to $E$ and is
therefore nonbinding; removals are determined by the two constraints above,
without a fitted transfer-cost threshold.

\section{Experiments}
\label{sec:experimental_setup}

\paragraph{Models and tasks.}
We evaluate three target--draft model pairs: Qwen3-30B-A3B-Instruct-2507,
Qwen3-Coder-30B-A3B-Instruct, and GPT-OSS-120B, each with its corresponding
EAGLE-3 draft model. We generate one greedy completion for every example in
the full evaluation sets of GSM8K~\citep{cobbe2021gsm8k} (flexible exact
match), MATH500~\citep{lightman2023lets} (\texttt{math\_verify}),
HumanEval~\citep{chen2021codex}, and MBPP~\citep{austin2021program} (pass@1
for both code benchmarks). For GPT-OSS on GSM8K and MATH500, we extract and
score the final Harmony channel.
We report full-residency accuracy and accuracy under residency-aware pruning
with $X=48$ slots per MoE layer.

\paragraph{Baselines.}
All speculative methods use the same EAGLE-3 draft tree with five draft steps,
an EAGLE top-$k$ of 10, and 64 draft tokens, isolating verifier-side expert
selection. The baselines are target-only autoregressive decoding (Vanilla AR),
Standard SD with the target model's native top-$k$ routing, and two
fixed-budget verifier-side selectors at the same $B_0$: Count-$B$, which keeps
the experts receiving the most natural top-$k$ votes in the block, and
\moespec, which ranks experts by router mass summed over the verification
block and retains the top $B_0$ experts~\citep{mcdanel2026moe}.

For each model--task pair, we compute the mean \sys expert-set size on the
first 50 evaluation prompts and round it to the nearest integer, with half
values rounded up. If the bootstrap interval crosses a rounding boundary, we
instead use the first 100 prompts. This budget-selection procedure uses only
routing and selector traces, never correctness outcomes, and $B_0$ is fixed
before either fixed-budget method is evaluated.
\sysfix is our matched-budget variant, which uses the same commitment-weighted
utility and root anchor as \sys but holds the expert-set size at $B_0$.
At the same budget, \moespec and \sysfix directly compare expert-selection
policies. Comparing \sysfix with \sys contrasts fixed-budget and self-sizing
selection, with $B_0$ matched to the rounded mean capacity selected by \sys.
Commitment probabilities are estimated
from disjoint training traces and held fixed during evaluation.

\paragraph{Throughput measurement.}
Full-residency throughput is measured at batch size one on one NVIDIA RTX PRO
6000 Blackwell using SGLang 0.5.12.post1~\citep{zheng2024sglang}, synchronous
Spec V1, overlap scheduling disabled, and CUDA Graphs enabled. After one
untimed warm-up request using evaluation example 50, we time examples 0--49 in
fixed order, generating exactly 256 output tokens per prompt while ignoring
EOS. We run three repetitions from fresh server instances, rotate the method
order across repetitions, and sum output tokens and timed generation across
the three repetitions to compute throughput for each model--task pair.

Physical expert-offloading throughput is measured for all three targets on one NVIDIA RTX 5090.
At $X=48$, all runs fit within its 32\,GB memory. GPT-OSS has the largest expert-pool
footprint and uses 28.60\,GiB at peak. Each configuration uses batch
size one, 20 frozen prompts, one untimed warm-up request, greedy decoding with
EOS ignored, and exactly 256 generated tokens per prompt. We report the median
of three fresh-server repetitions. The offload runtime uses SGLang
0.5.12.post1 with overlap scheduling and CUDA Graphs disabled because the
physical slot cache requires eager execution. Non-expert target weights, the draft model,
the KV cache, and the per-layer slot pools remain on the GPU. Qwen expert weights are stored in pinned host memory in BF16, whereas
GPT-OSS expert weights retain their native packed-MXFP4 representation in
pinned host memory. Load-time redirection
ensures that the complete checkpoint expert stacks are never simultaneously
loaded into GPU memory. Host-to-device (H2D) traffic and cache-hit statistics
are collected in separate single-repetition instrumented runs over the same
prompts and are excluded from throughput aggregation.

\subsection{Accuracy}
\label{sec:results_accuracy}

\begin{table*}[t]
\centering
\footnotesize
\setlength{\tabcolsep}{3.4pt}
\renewcommand{\arraystretch}{1.12}
\caption{Task accuracy (\%) for Qwen3-Instruct (Inst.), Qwen3-Coder (Coder),
and GPT-OSS-120B (OSS). $B_0$ is the rounded mean number of experts selected
by \sys on each model--task pair and is supplied to the fixed-budget methods.
$\tau$ is the mean accepted length per target pass. Bold marks the best result
per column among the six full-residency rows. The final Acc. column is the
unweighted mean over the 12 pairs, and vs.\ AR is its relative change from
Vanilla AR.}
\label{tab:accuracy}
\begin{tabular}{lccc ccc ccc ccc cc}
\toprule
& \multicolumn{3}{c}{GSM8K} & \multicolumn{3}{c}{MATH500} &
\multicolumn{3}{c}{HumanEval} & \multicolumn{3}{c}{MBPP} &
\multicolumn{2}{c}{Average} \\
\cmidrule(lr){2-4}\cmidrule(lr){5-7}\cmidrule(lr){8-10}\cmidrule(lr){11-13}\cmidrule(lr){14-15}
Method & Inst. & Coder & OSS & Inst. & Coder & OSS & Inst. & Coder & OSS &
Inst. & Coder & OSS & Acc. & vs.\ AR \\
\midrule
Expert budget $B_0$ & 34 & 28 & 24 & 32 & 26 & 23 & 31 & 33 & 25 & 30 & 33 & 24 & -- & -- \\
$\tau$, Standard SD & 4.04 & 3.52 & 4.73 & 4.62 & 4.06 & 5.04 & 3.82 & 4.77 & 4.32 & 4.36 & 4.80 & 4.58 & -- & -- \\
$\tau$, \moespec & 3.96 & 3.52 & 4.84 & 4.66 & 3.94 & 5.06 & 3.86 & 4.77 & 4.31 & 4.24 & 4.84 & 4.56 & -- & -- \\
$\tau$, \sys & 3.97 & 3.55 & 4.75 & 4.66 & 4.07 & 5.04 & 3.81 & 4.74 & 4.34 & 4.40 & 4.82 & 4.55 & -- & -- \\
\midrule
\band{15}{Natural routing} \\
Vanilla AR & \textbf{95.98} & 94.24 & 90.45 & 95.80 & \textbf{88.00} & 93.20 & \textbf{93.29} & \textbf{93.29} & \textbf{94.51} & 78.60 & 78.60 & \textbf{96.20} & \textbf{91.01} & -- \\
Standard SD & 95.53 & 93.93 & 91.43 & 95.60 & 87.40 & 93.80 & \textbf{93.29} & \textbf{93.29} & 92.68 & 79.60 & \textbf{78.80} & 95.40 & 90.90 & \delt{$-0.1\%$} \\
\midrule
\band{15}{Fixed budget $B_0$} \\
Count-$B$ & 90.98 & 91.51 & 90.90 & 94.80 & 82.00 & 93.00 & 76.83 & 63.41 & 90.24 & 67.20 & 61.20 & 90.60 & 82.72 & \delt{$-9.1\%$} \\
\moespec & 94.77 & 92.72 & \textbf{91.89} & 96.00 & 84.60 & \textbf{94.00} & 89.63 & 80.49 & 91.46 & 76.40 & 71.40 & 94.40 & 88.15 & \delt{$-3.1\%$} \\
\oursrow \sysfix & \textbf{95.98} & 94.16 & 91.05 & \textbf{97.00} & 87.00 & \textbf{94.00} & 89.63 & 92.68 & \textbf{94.51} & \textbf{80.00} & 76.60 & 94.60 & 90.60 & \delt{$-0.5\%$} \\
\midrule
\band{15}{Self-sizing expert set} \\
\oursrow \sys & 95.53 & \textbf{94.47} & 91.28 & 96.00 & 86.80 & 93.80 & 92.07 & 92.68 & 92.68 & 79.40 & 78.20 & 94.60 & 90.63 & \delt{$-0.4\%$} \\
\oursrow \sys (Offload) & 95.38 & 93.78 & 91.28 & 95.60 & 87.40 & 93.80 & 93.29 & 93.29 & 90.24 & 79.00 & 76.60 & 93.40 & 90.25 & -- \\
\bottomrule
\end{tabular}
\end{table*}

Table~\ref{tab:accuracy} separates the two decisions a verifier-side selector
makes. The \emph{Fixed budget} regime holds the number of selected experts at
$B_0$ and changes only the expert-selection policy. The \emph{Self-sizing}
regime lets \sys determine size per verification block.

Count-$B$ and \moespec require the deployment to supply $B$. The pair-specific
budget $B_0$ spans $23$ to $34$ experts and equals the rounded mean cardinality
selected by \sys on each pair. Supplying this value to the fixed-budget
baselines matches their expert-set size to \sys for the same workload instead
of forcing one global $B$ across all pairs. Thus, $B_0$ provides a size-matched
comparison and is not tuned to maximize a baseline's accuracy.

At this matched budget, expert membership still affects accuracy.
Ranking by vote count reaches $82.72\%$ on average and \moespec reaches
$88.15\%$, compared with $90.60\%$ for \sysfix. Aggregate router mass gives
every draft node equal weight even though at most one root-to-leaf path is
accepted. Within the $B_0$-expert shortlist, demand from discarded branches
can therefore displace experts used by committed tokens. Commitment weighting
reduces the contribution of positions that are less likely to reach the
output. Because cardinality is fixed, the observed differences reflect the
membership construction rule rather than the number of selected experts. The
two selectors also have similar acceptance behavior. The largest
absolute difference in $\tau$ between \sys and \moespec is $0.17$ tokens, and
their signed median difference is below $0.01$ tokens. Both selectors remain
close to Standard SD, so the accuracy difference is not accompanied by a
material change in mean accepted length.

With self-sizing, \sys determines expert membership and cardinality per block.
Its mean accuracy over the 12 pairs is
$0.27$ points below Standard SD, with a mean absolute per-pair difference of $0.43$
points and a maximum of $1.22$ points. Compared with self-sizing, fixing the size
at the matched $B_0$ yields a mean absolute per-pair difference of $0.74$
points and a maximum of $2.44$ points. The self-sizing and fixed-budget
variants have nearly identical mean accuracy across the 12 pairs but differ on
individual pairs.

Under expert offloading, \sys uses an expert pool of $X=48$ slots per MoE layer
and modifies expert eligibility according to the current expert-cache state.
To measure the accuracy effect of this cache-aware optimization, we compare it
with the same selector with residency-aware pruning disabled on the same
offload stack. Enabling pruning changes mean accuracy by $-0.28$ points across
the eight Qwen pairs and $-0.27$ points across the four GPT-OSS pairs, with
per-pair changes from $-1.83$ to $+1.22$ points. Thus, cache-aware pruning
incurs less than $0.3$ percentage points of mean accuracy loss.

\subsection{End-to-End Throughput}

\begin{figure}[t]
\centering
\includegraphics[width=\columnwidth]{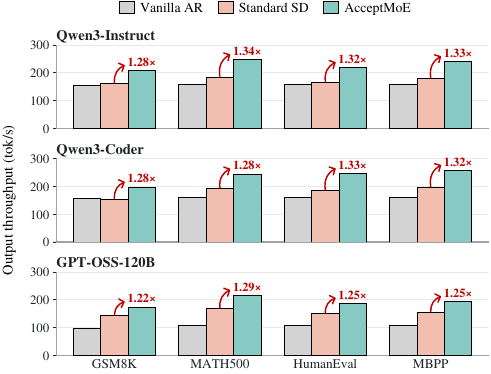}
\caption{End-to-end decoding throughput with all expert weights resident in
GPU memory, on one RTX PRO 6000 Blackwell.}
\label{fig:hbm-speed}
\end{figure}

Figure~\ref{fig:hbm-speed} reports end-to-end throughput. \sys outperforms
Standard SD on all 12 model--task pairs, with a mean speedup of $1.290\times$
and a per-pair range of $1.217$--$1.339\times$. \sys and Standard SD both have
a mean accepted length of $4.39$ tokens after rounding. Their largest per-pair
difference is $0.07$ tokens. The throughput improvement therefore does not
come from longer accepted prefixes. The result is consistent with smaller
eligible sets reducing per-pass expert computation and expert-weight accesses.
Combining output tokens and timed generation across the four tasks for each
model gives per-model speedups of $1.249$--$1.318\times$. The gain is smaller
on GPT-OSS, whose lower per-token routing fan-out leaves a smaller natural
expert union and less scope for reduction. Model architecture and execution
kernels also differ across targets, so this comparison does not isolate the
effect of routing fan-out.

\begin{figure}[t]
\centering
\includegraphics[width=\columnwidth]{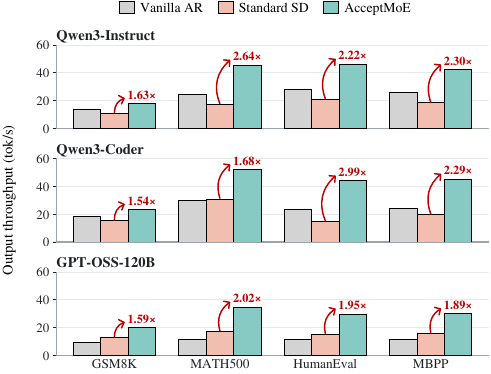}
\caption{End-to-end decoding throughput under physical expert offloading, on
one RTX 5090 with 48 expert slots per MoE layer.}
\label{fig:offload-speed}
\end{figure}

\begin{figure*}[t]
\centering
\includegraphics[width=\textwidth]{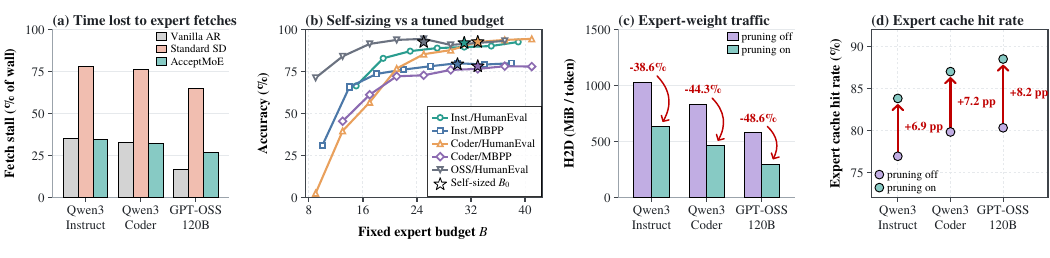}
\caption{Offload mechanism and ablations. (a) Fraction of wall time spent
waiting for expert fetches. (b) Accuracy across a sweep of fixed expert
budgets. (c) Expert-weight traffic and (d) expert-cache hit rate with
residency-aware pruning disabled and enabled.}
\label{fig:ablations}
\end{figure*}

Expert offloading exposes the cost of the verifier-side expert union. Standard
SD applies natural routing to every node in the draft tree, so each verification
pass may request the union of experts used across the entire tree. This expanded
union lowers cache reuse and increases expert-weight transfers. The resulting
transfer overhead can outweigh the benefit of verifying tokens in parallel.
Standard SD spends $65\%$ to $78\%$ of wall time waiting for expert weights and
is slower than Vanilla AR on seven of the eight Qwen pairs. \sys restricts the
eligible expert union for each verification block and reduces the fetch-stall
share to $27\%$--$35\%$. It outperforms both Standard SD and Vanilla AR on all
12 model--task pairs and achieves a mean speedup of $2.06\times$ over Standard
SD. Figure~\ref{fig:offload-speed} reports the throughput results, and
Figure~\ref{fig:ablations} reports the wait-time shares. Standard SD remains
faster than Vanilla AR on all four GPT-OSS pairs. This difference is consistent
with GPT-OSS expert weights using packed MXFP4 rather than BF16.

\begin{table}[t]
\centering
\footnotesize
\setlength{\tabcolsep}{1.4pt}
\renewcommand{\arraystretch}{1.12}
\caption{Expert-weight traffic and cache behavior under expert offloading. For
each model, H2D traffic and cache hit rate are computed after summing their
counters across four tasks. H2D is host-to-device expert traffic per output
token. $\Delta$ denotes the percentage change in H2D traffic and the
percentage-point change in hit rate. Qwen expert weights are transferred in
BF16, and GPT-OSS expert weights in packed MXFP4. H2D byte counts are therefore
not directly comparable across targets.}
\label{tab:offload-transfer}
\begin{tabular}{lrrr@{\hspace{5pt}}rrr}
\toprule
& \multicolumn{3}{c}{H2D (MiB/tok) $\downarrow$} &
\multicolumn{3}{c}{Hit rate (\%) $\uparrow$} \\
\cmidrule(lr){2-4}\cmidrule(lr){5-7}
Method & Inst. & Coder & OSS & Inst. & Coder & OSS \\
\midrule
Vanilla AR & 888 & 721 & 439 & 81.0 & 83.5 & 81.0 \\
Standard SD & 2394 & 2025 & 1243 & 73.0 & 74.4 & 74.3 \\
\oursrow \sys & \textbf{633} & \textbf{464} & \textbf{298} & \textbf{83.8} & \textbf{87.0} & \textbf{88.5} \\
\addlinespace[1pt]
$\Delta$ & \gain{\scriptsize $-73.6\%$} & \gain{\scriptsize $-77.1\%$} &
\gain{\scriptsize $-76.0\%$} &
\gain{\scriptsize $+10.8$} & \gain{\scriptsize $+12.6$} &
\gain{\scriptsize $+14.2$} \\
\bottomrule
\end{tabular}
\end{table}

Table~\ref{tab:offload-transfer} quantifies this mechanism. Standard SD
transfers more expert-weight data than Vanilla AR on all three targets and has
a lower cache hit rate. Compared with Standard SD, \sys reduces expert-weight
traffic per output token by $73.6$--$77.1\%$ and raises the expert-cache hit
rate by $10.8$ to $14.2$ percentage points. Together, these measurements show
that \sys reduces offload overhead by transferring fewer expert weights and
reusing cached experts more effectively.

\subsection{Ablation Study}
\label{sec:results_ablation}
We ablate two components of \sys's verifier-side expert selection: expert-set
cardinality and residency-aware pruning.

\paragraph{Self-sizing versus measured fixed budgets.}
The fixed-budget sweep in Figure~\ref{fig:ablations} evaluates five
model--task pairs while holding the membership rule fixed. The self-sized point
is $0.97$ percentage points below the best measured fixed-budget point in its
own sweep on average and $1.83$ points below it in the worst case. At the
smallest displayed budgets, losses compared with each sweep's best measured point
range from $23.2$ to $92.1$ points. A budget selected for one workload also need
not transfer. On HumanEval, $B{=}25$ is the best measured budget for
GPT-OSS-120B. Applying this budget to Qwen3-Coder on the same task yields
accuracy $9.1$ percentage points below Qwen3-Coder's best measured fixed-budget
result. Across these five sweeps,
self-sizing avoids a per-pair budget sweep with an average measured accuracy gap
of $0.97$ percentage points.

\paragraph{Residency-aware pruning.}
To isolate residency awareness, we compare \sys with a variant that disables
residency-aware pruning. Both configurations construct the same
commitment-weighted, self-sized expert set before pruning. The disabled variant
leaves this set unchanged and does not use cache residency to modify expert
eligibility. Figure~\ref{fig:ablations} reports the traffic and cache results.
Enabling residency-aware pruning reduces expert-weight traffic by
$38.6\%$ to $48.6\%$ and raises cache hit rate by $6.9$ to $8.2$ percentage
points. It also improves throughput by $4.6\%$ to $15.1\%$. Enabling pruning
changes mean accuracy by $-0.27$ points across the 12 pairs, with per-pair
changes from $-1.83$ to $+1.22$ points.
The largest throughput gain and traffic reduction both occur on GPT-OSS.
These results show that configurations with similar mean accuracy across the
12 pairs can nevertheless differ substantially in transfer traffic.


\section{Conclusion}

We presented \sys, a verifier-side expert selector for speculative decoding
with MoE targets. Verification cost depends not only on the number of verified
tokens but also on the union of experts they activate. Under expert offloading,
it also depends on which experts are nonresident. \sys combines target-router
scores with offline-estimated commitment probabilities and automatically
adjusts the number of eligible experts for each verification block. This
self-sizing design removes the need for a predefined expert budget. Under
partial residency, \sys further conditions expert eligibility on cache
residency to reduce transfers without a learned expert predictor.

Across three targets and four tasks, \sys's mean accuracy is $0.27$ percentage
points lower than that of Standard SD. Compared with Standard SD, it reaches
$1.290\times$ the throughput with all experts resident and $2.06\times$ under
expert offloading. It also reduces host-to-device traffic by $73.6\%$ to
$77.1\%$. These results show that verifier-side expert selection benefits from
adapting expert-set size to each verification block and expert eligibility to
the current cache state.

\clearpage
\bibliography{aaai2027}

\end{document}